\documentclass{article}

\usepackage[nonatbib,preprint]{neurips_2020}

\usepackage[pagebackref=true,breaklinks=true,colorlinks,bookmarks=false,citecolor=blue,linkcolor=blue]{hyperref}       
\usepackage[utf8]{inputenc} 
\usepackage[T1]{fontenc}    
\usepackage{url}            
\usepackage{booktabs}       
\usepackage{amsfonts}       
\usepackage{nicefrac}       
\usepackage{microtype}      
\usepackage{amsmath,amssymb} 
\usepackage{color}
\usepackage{multirow}
\usepackage{subfigure}
\usepackage{caption}
\usepackage{graphicx}
\usepackage{enumitem}
\usepackage{pifont}
\setitemize{noitemsep,topsep=0pt,parsep=0pt,partopsep=0pt}
\newcommand{\etal}{\textit{et al. }}
\newcommand{\cmark}{\text{\ding{51}}}
\newcommand{\xmark}{\text{\ding{55}}}

\title{Real-time Universal Style Transfer on High-resolution Images via Zero-channel Pruning}

\author{%
  Jie An\thanks{This work is done when Jie An works as an intern at Tencent AI Lab.} \\
  Department of Computer Science\\
  University of Rochester\\
  \texttt{jan6@cs.rochester.edu} \\
  \And
    Tao Li\thanks{Equal contribution.} \\
  School of Mathematical Sciences\\
  Peking University\\
  \texttt{li\_tao@pku.edu.cn} \\
   \And
    Haozhi Huang\\
  AI Lab\\
  Tencent Inc.\\
  \texttt{matthuang@tencent.com} \\
  \And
    Li Shen\\
  AI Lab\\
  Tencent Inc.\\
  \texttt{mathshenli@gmail.com} \\
  \And
    Xuan Wang\\
  AI Lab\\
  Tencent Inc.\\
  \texttt{xwang.cv@gmail.com} \\
  \And
    Yongyi Tang\\
  AI Lab\\
  Tencent Inc.\\
  \texttt{yongyi.tang92@gmail.com} \\
  \And
    Jinwen Ma\\
  School of Mathematical Sciences\\
  Peking University\\
  \texttt{jwma@math.pku.edu.cn} \\
  \And
    Wei Liu\\
  AI Lab\\
  Tencent Inc.\\
  \texttt{wl2223@columbia.edu} \\
  \And
    Jiebo Luo\\
  Department of Computer Science\\
  University of Rochester\\
  \texttt{jluo@cs.rochester.edu} \\
}

\begin{document}

\maketitle

\begin{abstract}
    Extracting effective deep features to represent content and style information is the key to universal style transfer. Most existing algorithms use VGG19 as the feature extractor, which incurs a high computational cost and impedes real-time style transfer on high-resolution images. In this work, we propose a lightweight alternative architecture -- \emph{ArtNet}, which is based on GoogLeNet, and later pruned by a novel channel pruning method named \emph{Zero-channel Pruning} specially designed for style transfer approaches. Besides, we propose a theoretically sound \emph{sandwich swap transform} ($\mathbf{S}^2$) module to transfer deep features, which can create a pleasing holistic appearance and fine local textures with an improved content preservation ability. By using ArtNet and $\mathbf{S}^2$, our method is $2.3\sim 107.4\times$ faster than state-of-the-art approaches. The comprehensive experiments demonstrate that ArtNet can achieve universal, real-time, and high-quality style transfer on high-resolution images simultaneously (68.03 FPS on $512\times512$ images). 
\end{abstract}

\vspace{-2mm}
\section{Introduction}
\begin{figure}[t]
    \centering
    \includegraphics[width=\textwidth]{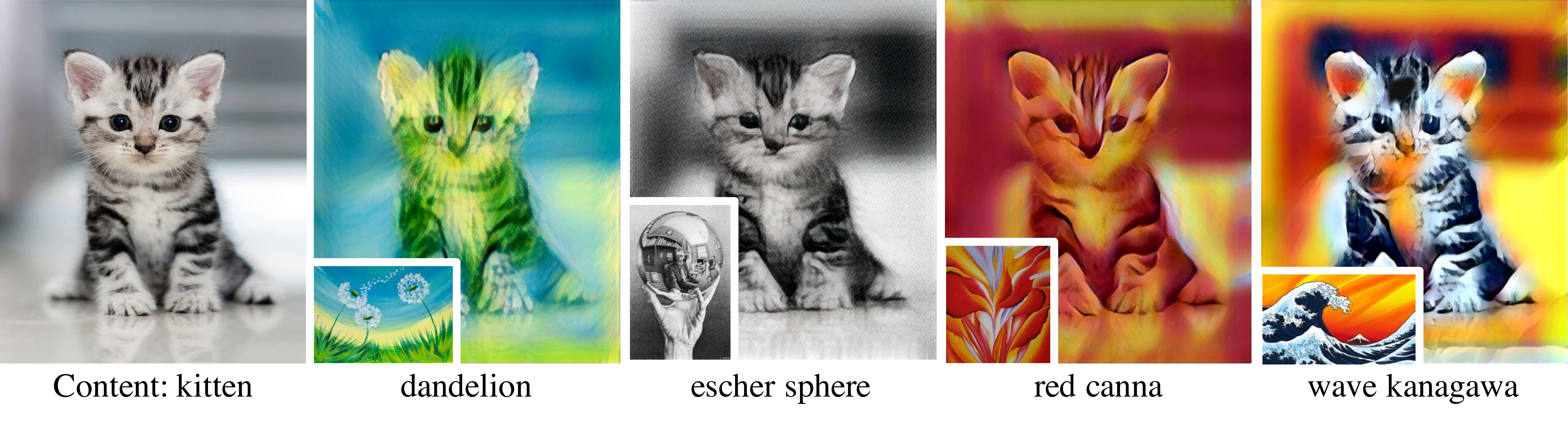}
    \captionof{figure}{\small Style transfer results with our algorithm on arbitrary content and style. Our algorithm achieves 68.03 FPS on $512 \times 512$ images.}
    \vspace{-5mm}
    \label{fig:1}
\end{figure}

\vspace{-2mm}
Neural style transfer is an image editing task that aims at changing the artistic style of an image according to a reference image. Given a pair of content and style images as the input, a style transfer method will generate an image with the scene of the content and the visual effects ($e.g.$, colors, textures, strokes, etc.) of the style image. For example, in the top row of Fig.~\ref{fig:1}, we transfer a picture of ``kitten'' into different styles. In the bottom row, the produced images preserve the scene content of the ``Chureito pagoda'' but inherit the visual styles from several reference photos. 

Remarkable advances have been made in neural style transfer. The pioneering work is presented in~\cite{gatys2015texture,Gatys2016}, where Gatys~\etal make the first attempt to connect style representation to the Gram matrices of extracted deep features. 
Their work shows that Gram matrices of deep features of a VGG19 network have an exceptional ability to encode images' styles.
Following this line of research, several iterative algorithms that aim to minimize a Gram-based loss function~\cite{gatys2016preserving,risser2017stable,luan2017deep,li2017laplacian,li2017demystifying,huang2017real} have been proposed. While these algorithms can produce high-quality stylization results, they all suffer from a high computational cost since the optimization method they used usually needs hundreds of iterations. 
By discarding time-consuming iterations, a few style transfer methods based on feed-forward neural networks~\cite{frigo2016split,ulyanov1607instance,johnson2016perceptual,ulyanov2016texture,risser2017stable,ulyanov2017improved,dumoulin2017learned,chen2017stylebank,zhang2017multi,wang2017multimodal,gong2018neural} have been proposed. While these algorithms can produce style transfer results with improved efficiency, they usually can only work well on a limited number of image styles. Therefore, the practical application of these methods is limited.
Recently, universal style transfer methods~\cite{chen2016fast,li2017universal,huang2017arbitrary,sheng2018avatar,li2018closed,gu2018arbitrary,yoo2019photorealistic,an2019ultrafast,jing2019dynamic,li2018learning,lu2019optimal} have been proposed for handling \emph{arbitrary} styles and contents. To generate high-quality images while enjoying the benefit of universal transfer, a few improvements such as multi-level stylization~\cite{li2017universal,li2018closed,li2018learning,lu2019optimal}, iterative EM process~\cite{gu2018arbitrary}, and wavelet transform~\cite{yoo2019photorealistic} have been introduced. However, the above improvements also come with a high computation cost.
It is therefore desirable to design a novel approach to achieve universal ($i.e.$, arbitrary), fast ($e.g.$, real-time on high-resolution images), and high-quality style transfer at the same time. Our work fills this gap.

VGG19~\cite{simonyan2014very} has been used by most style transfer algorithms. For example, iterative methods~\cite{Gatys2016,gatys2016preserving,risser2017stable,luan2017deep,li2017laplacian,li2017demystifying} use VGG19 as the feature extractor and compute loss terms accordingly. On the other hand, approaches based on feed-forward neural networks~\cite{li2017universal,li2018closed,gu2018arbitrary,yoo2019photorealistic} adopt VGG19 to form auto-encoders. VGG19 can indeed produce high-quality style transfer images, but has a serious drawback: 
VGG19 has 12.95 MB parameters and 189.50 GFLOPs, which imposes a huge computational burden (incurring 0.04s for a 512 $\times$ 512 image on an NVIDIA TitanXp GPU card) and makes real-time style transfer on high-resolution images completely impossible. 

To overcome the drawback of the VGG19 and achieve real-time style transfer, we propose an alternative architecture. The proposed architecture is based on GoogLeNet~\cite{szegedy2015going}, which is selected among 28 most popular neural network architectures~\cite{simonyan2014very,krizhevsky2014one,iandola2016squeezenet,huang2017densely,szegedy2016rethinking,szegedy2015going,ma2018shufflenet,sandler2018mobilenetv2,xie2017aggregated,zagoruyko2016wide} in terms of stylization effect and efficiency. 

Instead of directly using GoogLeNet, we propose a new channel pruning method (named \emph{Zero-channel Pruning}) to further reduce the computational cost by removing redundant channels of deep features. The Zero-channel pruning algorithm is motivated by our finding that deep features by the ReLU layers in the networks ($e.g.$, GoogLeNet) pre-trained on the ImageNet dataset~\cite{krizhevsky2012ImageNet} contain a few empty channels. We call these blank feature maps \emph{Zero Channels}. The proposed algorithm prunes the network by removing those Zero Channels and the corresponding channels in Convolution and BN operators ahead of the ReLU layer in the same position. The pruned GoogLeNet (named ArtNet) achieves 2$\times$ acceleration at the cost of little performance degradation. Compared to~\cite{li2016pruning}, the proposed Zero-channel pruning method adopts a different channel selection criterion and can achieve degradation-free acceleration in style transfer without any fine-tuning process.
Thanks to the proposed Zero-channel pruning method, ArtNet can extract features with up to $16\times$ down-sampling instead of compromising to $4\times$~\cite{chen2016fast} or $8\times$~\cite{huang2017arbitrary}. Therefore, our network architecture has a much larger field-of-view and can capture more complicated local structures and textures~\cite{li2017universal}.

Besides, we address one fundamental problem in style transfer by introducing a theoretically sound correction in the feature transfer modules. We propose a \emph{sandwich swap transform} ($\mathbf{S}^2$) module to replace those transfer modules ($e.g.$, StyleSwap~\cite{chen2016fast}, WCT~\cite{li2017universal}, AdaIN~\cite{huang2017arbitrary}, and Avatar-Net~\cite{sheng2018avatar}) at the bottlenecks of auto-encoders. Our motivation is to take advantage of AdaIN, which excels in creating pleasing global appearance while preserving the textures/paint strokes transfer of StyleSwap. The proposed $\mathbf{S}^2$ module, with improved content preservation, enables our algorithm to achieve exact coloring matching, complex textures reproduction, and global appearance transfer at the same time.

Our main contributions can be summarized as follows:
\begin{itemize}
    \item We propose a \emph{Zero-channel Pruning} algorithm specially designed for style transfer. The algorithm can remove the network's redundant parameters while keeping the stylization quality almost intact even without a fine-tuning process.
    \item We propose ArtNet -- a compact model for real-time style transfer on high-resolution images. ArtNet is $2.3\sim 107.4\times$ faster than the state-of-the-art algorithms and can achieve a real-time speed of 68.03 FPS on 512 $\times$ 512 images.
    \item We propose a theoretically sound sandwich swap transform ($\mathbf{S}^2$) to provide improved content preservation, enabling the transfer of fine local details and global appearance simultaneously.
\end{itemize}
\vspace{-2mm}
\section{Related Work}
\vspace{-2mm}
\paragraph{Universal Style Transfer.}

Most universal style transfer algorithms~\cite{chen2016fast,huang2017arbitrary,li2017universal,li2018closed,gu2018arbitrary,sheng2018avatar,yoo2019photorealistic} consist of two parts: an auto-encoder and a style transfer module that works in the bottleneck. 
Among these algorithms, the auto-encoder architecture used by all these methods is the same, $i.e.$, the VGG19 network~\cite{simonyan2014very} pre-trained on the ImageNet dataset~\cite{krizhevsky2012ImageNet} as the encoder and its inverted version as the decoder. One main difference between our algorithm and other universal style transfer approaches is that we adopt an auto-encoder with a new architecture. Such a new auto-encoder (ArtNet) has one third of the parameters and is the first architecture that enables real-time, universal, high-quality style transfer on high-resolution images simultaneously.

StyleSwap~\cite{chen2016fast}, AdaIN~\cite{huang2017arbitrary}, and Avatar-Net~\cite{sheng2018avatar} are the most relevant approaches to ours. Besides a different auto-encoder architecture, our method also differs importantly from the previous methods in terms of a new transfer module.
The proposed $\mathbf{S}^2$ module performs feature transfer as \emph{AdaIN-Swap-AdaIN}, in which \emph{Swap}~\cite{chen2016fast} is used to handle sophisticated textures/strokes directly while \emph{AdaIN}~\cite{huang2017arbitrary} is employed to transfer holistic global appearance. The combination of these two modules outperforms using each module individually. Compared to Avatar-Net~\cite{sheng2018avatar}, the proposed feature transfer mechanism ($i.e.$, AdaIN-Swap-AdaIN) is more theoretically sound and performs faster by avoiding the use of the ``hourglass'' strategy.

\vspace{-2mm}
\paragraph{Image-to-Image Translation.}
In addition to style transfer, image-to-image translation~\cite{isola2017image,wang2018high,liu2016coupled,taigman2016unsupervised,shrivastava2017learning,liu2017unsupervised,zhu2017unpaired,huang2018multimodal,liu2019few} can also be used to transfer image styles. To render a certain visual style, image-to-image translation methods usually need pre-transfer and post-transfer datasets to train generator and discriminator networks. However, universal style transfer methods ($e.g.$, our algorithm) 
can be used to make style transfer for arbitrary content and style images in a zero-shot fashion. 

\vspace{-2mm}
\paragraph{Neural Network Pruning.}
The network pruning approaches can be categorized into two types: weight pruning and channel pruning. Weight pruning algorithms~\cite{han2015learning,lee2018snip,carreira2018learning,ding2019global} usually detect non-operative weight positions in filters and disable them by setting to zero. While channel pruning approaches~\cite{li2016pruning,he2018soft,yu2018nisp,he2019filter} delete information-wise redundant channels and their corresponding weights entirely. The proposed zero-channel pruning method belongs to the channel pruning category. Since style transfer methods tend to use feature maps produced by ReLU layers in networks, our method removes Zero Channels of every ReLU layer and hereby indirectly deletes the weights of the preceding convolution/batch normalization operators. This differs from other general channel pruning methods that focus on pruning channels of convolutional layers directly. More importantly, the proposed pruning method does not need any fine-tuning process in making style transfer.

\begin{figure}[t]
    \centering
    \includegraphics[width=0.999\textwidth]{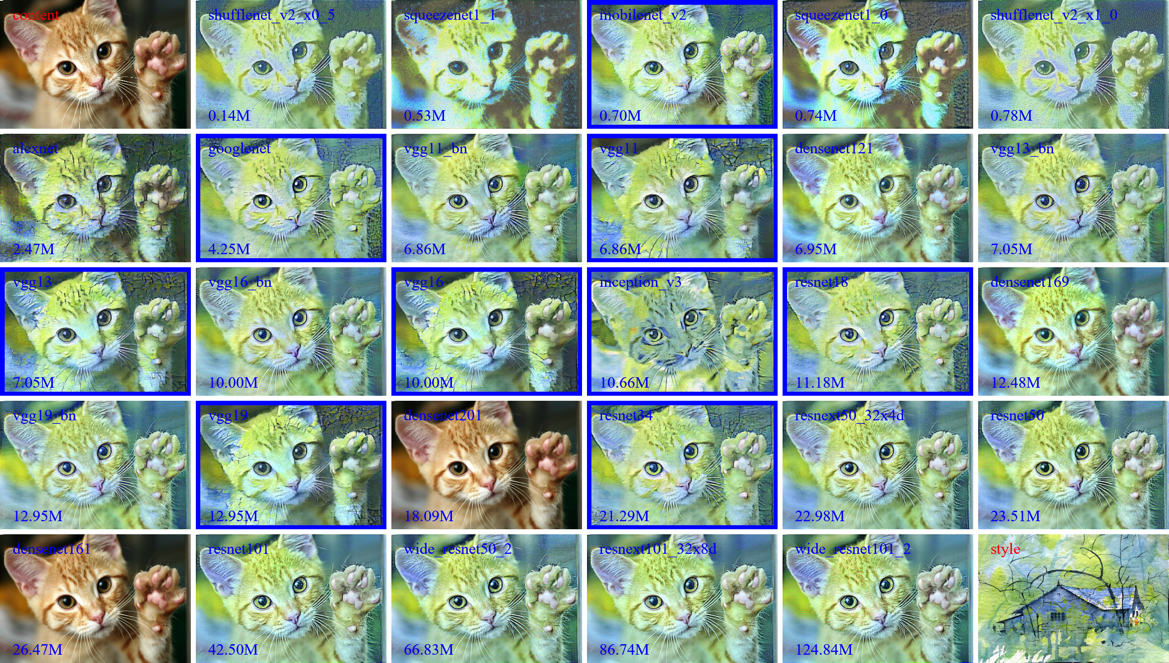}
    \vspace{-3mm}
    \caption{\small Style transfer results by 28 compared network architectures. In each subfigure, the architecture name is shown in the top-left corner while the parameter size is shown in the bottom-left corner. Among these networks, architectures highlighted by blue boxes outperform the others in rendering fine textures. Please zoom in on screen to appreciate fine details. We select GoogLeNet as the base model since it is the most compact architecture that can produce comparable style transfer results to VGG19 in terms of fine texture rendering.}
    \vspace{-3mm}
    \label{fig:2}
\end{figure}

\vspace{-2mm}
\section{Pre-Analysis}
\vspace{-2mm}
Existing state-of-the-art style transfer algorithms~\cite{li2017universal,li2018closed,gu2018arbitrary,yoo2019photorealistic} commonly use the VGG19 network~\cite{simonyan2014very} as the feature extractor.
The architecture of the feature extractor plays a central role in style transfer algorithms. On the one hand, deep features extracted by it directly influence the quality of style transfer. Concerning efficiency, feature extractors usually occupy a large portion of the overall time cost. 
Therefore, a more lightweight feature extractor can lead to a faster style transfer. 
To improve feature extractors for \emph{high-quality} style transfer on \emph{high-resolution} images at \emph{real-time} speed, it is necessary to make an in-depth analysis of the feature extractor. We start by answering the following three questions:

\vspace{-2mm}
\paragraph{Q1: Is VGG19 the One-Above-All?}
VGG19 is widely used by existing style transfer algorithms since Gatys~\etal\cite{gatys2015neural}. VGG19 can indeed produce superior style transfer results over other network architectures. However, VGG19 has to incur 0.04s for inference on a 512$\times$512 image. Therefore, \emph{style transfer algorithms using VGG19 as the feature extractor cannot reach real-time speed on high-resolution (more than 512$\times$512) images absolutely.} To make \emph{high-quality}, \emph{real-time} style transfer on \emph{high-resolution} images, we have to break the speed bottleneck by replacing VGG19 with other alternatives.

\vspace{-2mm}
\paragraph{Q2: Which architecture can be the alternative concerning the performance and efficiency?}
To find an efficient network architecture with a strong style transfer ability, we perform a comparison in terms of the style transfer performance and time consumption among 28 popular network architectures~\cite{simonyan2014very,krizhevsky2014one,iandola2016squeezenet,huang2017densely,szegedy2016rethinking,szegedy2015going,ma2018shufflenet,sandler2018mobilenetv2,xie2017aggregated,zagoruyko2016wide} upon a dataset that consists of 1092 content-style pairs (42 content; 26 style).
We first train each network on the ImageNet dataset~\cite{krizhevsky2012ImageNet}, and then use these networks to generate stylized images based on the algorithm of Gatys~\etal\cite{gatys2015neural}, respectively. Here we apply the algorithm proposed by Gatys~\etal\cite{gatys2015neural} since it does not need a decoder to invert features back to images, thus avoiding the bias introduced by the decoder training.
Fig.~\ref{fig:2} shows an example of the produced results of a content-style pair. 
According to the visual comparison, VGG19/16/13/11, ResNet18/34, GoogLeNet, Inceptionv3, and Mobilenetv2 (highlighted by blue boxes) outperform other architectures in terms of fine local texture generation. We then conduct a user study based on these architectures of good quality. In the user study, subjects are asked to rank the results of different methods. The mean rank result (GoogLeNet: 2.88, Inceptionv3: 2.95, Mobilenetv2: 2.91, ResNet34: 3.06, VGG19: 2.99, less is better) shows that: 1) VGG19 cannot outperform other networks significantly; 2) GoogLeNet can produce comparable style transfer results. 
However, GoogLeNet only contains $1/3$ of the parameters compared with VGG19. 
To balance the performance and efficiency, we choose GoogLeNet as the replacement of the VGG19. Please refer to the supplementary material for the user study details and more results.

\vspace{-2mm}
\paragraph{Q3: Can the potential alternative be further improved to make style transfer even faster?}
By employing GoogLeNet instead of VGG19 as the feature extractor, a style transfer algorithm ($e.g.$, Gatys~\cite{gatys2015neural} or AdaIN~\cite{huang2017arbitrary}) can achieve more than $3\times$ acceleration. However, the additional speed improvement can save more time budget for more complicated style transfer modules.
\begin{table}[t]
\vspace{-3mm}
    \caption{\small Style transfer methods comparison.}
\centering
    \scriptsize \setlength{\tabcolsep}{6.1pt}
    \begin{tabular}{lccccccccc}
        \toprule
        Method & Gatys~\etal & StyleSwap & AdaIN & WCT & LinearWCT & OptimalWCT & Avatar-Net & \textbf{ArtNet}\\
        \midrule
        Real-time & $\xmark$ & $\xmark$ & $\cmark$ & $\xmark$ & $\xmark$ & $\xmark$ & $\xmark$ & $\cmark$ \\
        Universal & $\xmark$ & $\cmark$ & $\cmark$ & $\cmark$ & $\cmark$ & $\cmark$ & $\cmark$ & $\cmark$  \\
        High-quality & $\cmark$ & $\xmark$ & $\xmark$ & $\cmark$ & $\cmark$ & $\cmark$ & $\cmark$ & $\cmark$ \\
        Content Preservation & $\xmark$ & $\cmark$ & $\xmark$ & $\xmark$ & $\xmark$ & $\xmark$ & $\xmark$ & $\cmark$ \\
        \bottomrule
    \end{tabular}
        \vspace{-5mm}
    \label{tab:2}
\end{table}

By visualizing the feature maps per channel, we find that a few channels (especially in shallower layers) in ReLU activation are completely empty. We call those empty channels \emph{Zero Channels}. Please refer to the supplementary material for feature map visualization and channel norm statistics on the MS\_COCO dataset, which quantitatively demonstrates the zero channel phenomenon. 
According to quantitative analysis on a large number of images, we find that positions of those Zero Channels are consistent among input images, $i.e$., are data agnostic. 
During inference, Zero Channels have no effect on the following layers (please refer to the supplementary material for the mathematical proof). However, Zero Channels themselves, the corresponding convolution and BN operators in upper layers still waste GPU memories and computational resources. Since only the features of ReLU layers are used to make style transfer, we can accelerate the style transfer algorithm without damaging the transfer quality by pruning those Zero Channels and other corresponding parameters in upper convolutional and BN layers.

\begin{figure}[t]
    \centering
    \includegraphics[width=0.8\textwidth]{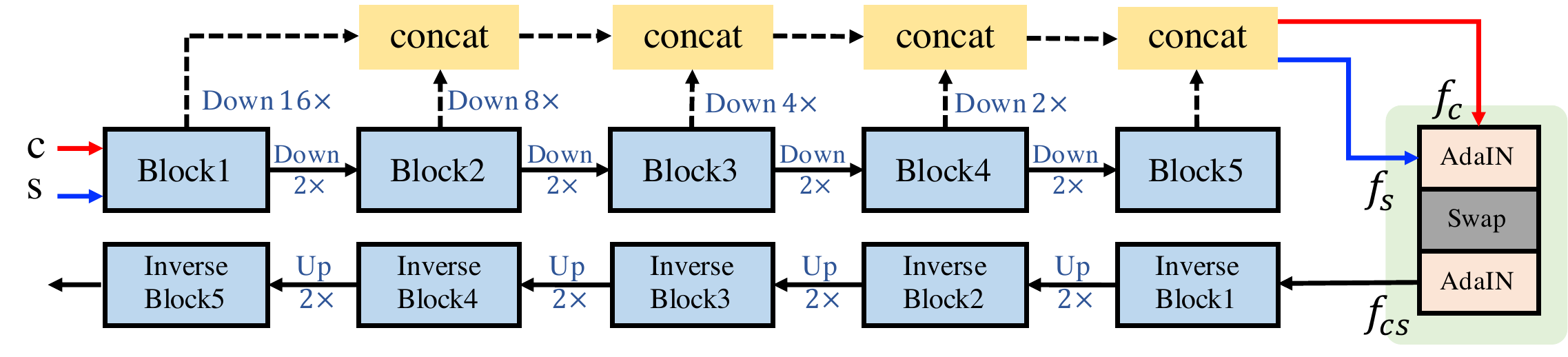}
    \vspace{-3mm}
    \caption{\small The framework of the proposed style transfer algorithm. Here Block/Inverse Block are network sections split by every pooling/upsampling layer, respectively. For ArtNet, these blocks are constructed by successive Inception modules.}
    \vspace{-5mm}
    \label{fig:6}
\end{figure}
\begin{figure}[t]
    \centering    
    \subfigure[\scriptsize{Prune Conv-BN-ReLU}] {
 \label{fig:4a}     
 \includegraphics[width=0.25\textwidth]{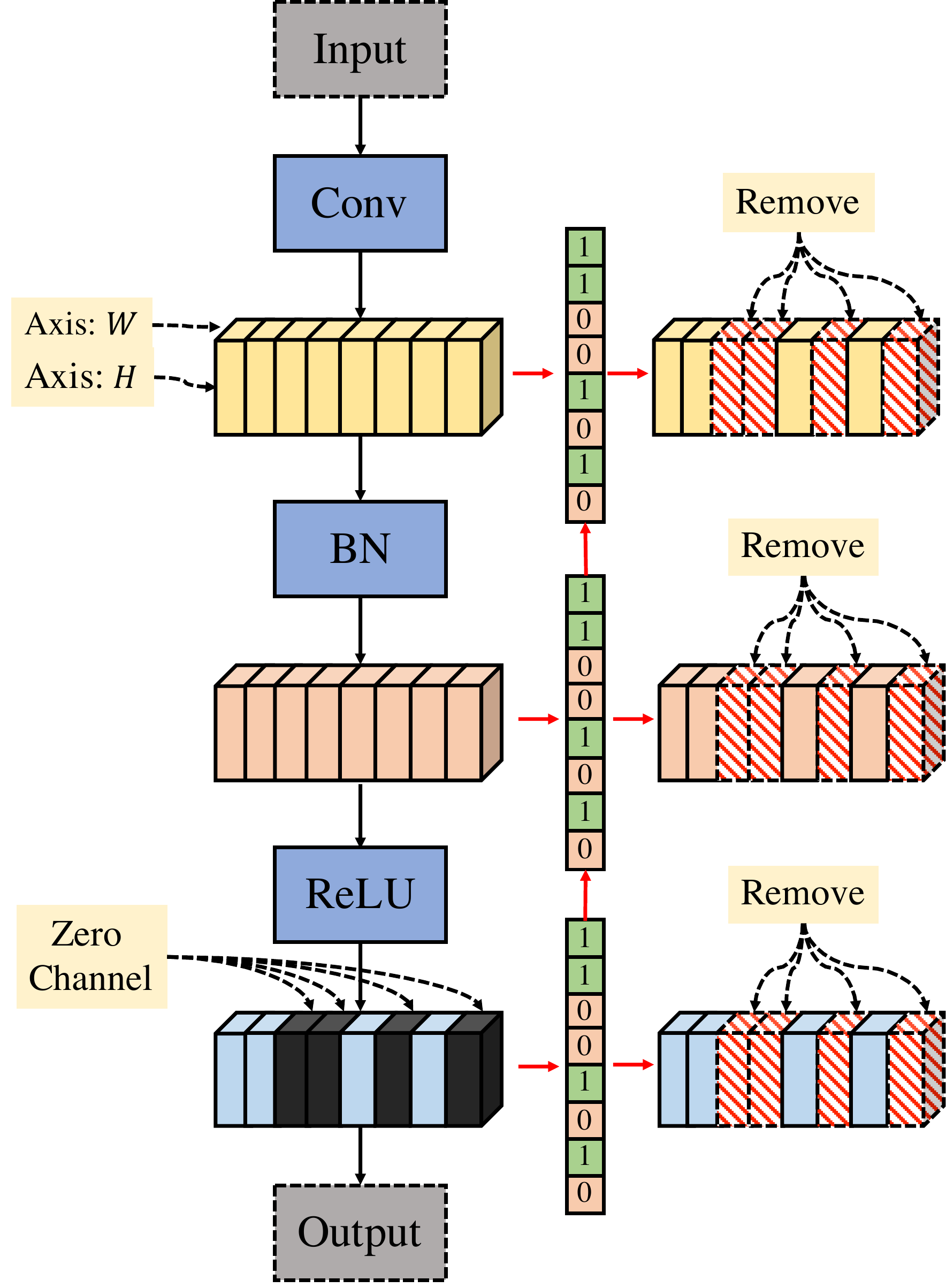}  
}      
\subfigure[\scriptsize{Prune Inception}] { 
\label{fig:4b}     
\includegraphics[width=0.41\textwidth]{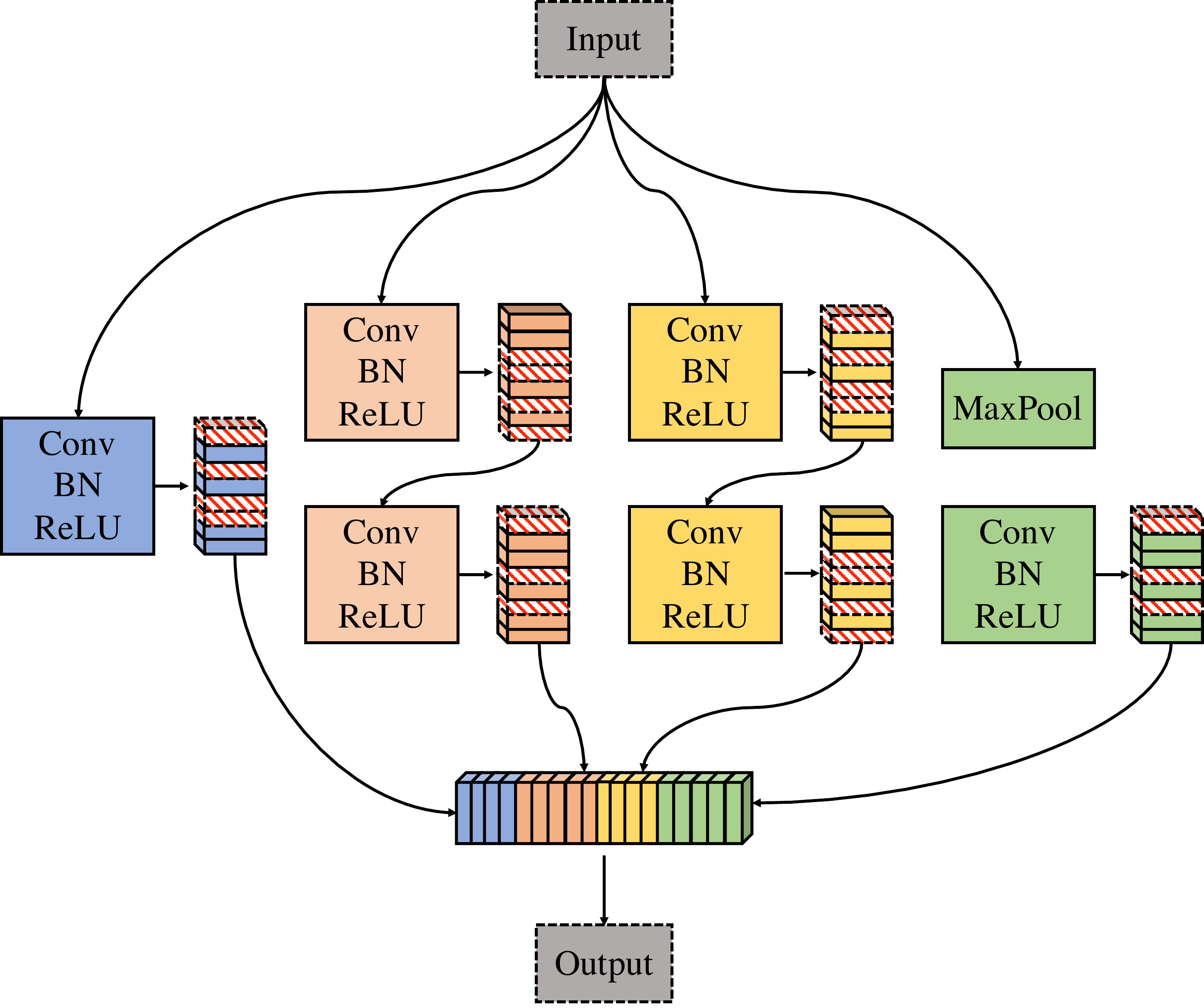}     
}    
\subfigure[\scriptsize{Prune InvertedRes}] { 
\label{fig:4c}     
\includegraphics[width=0.21\textwidth]{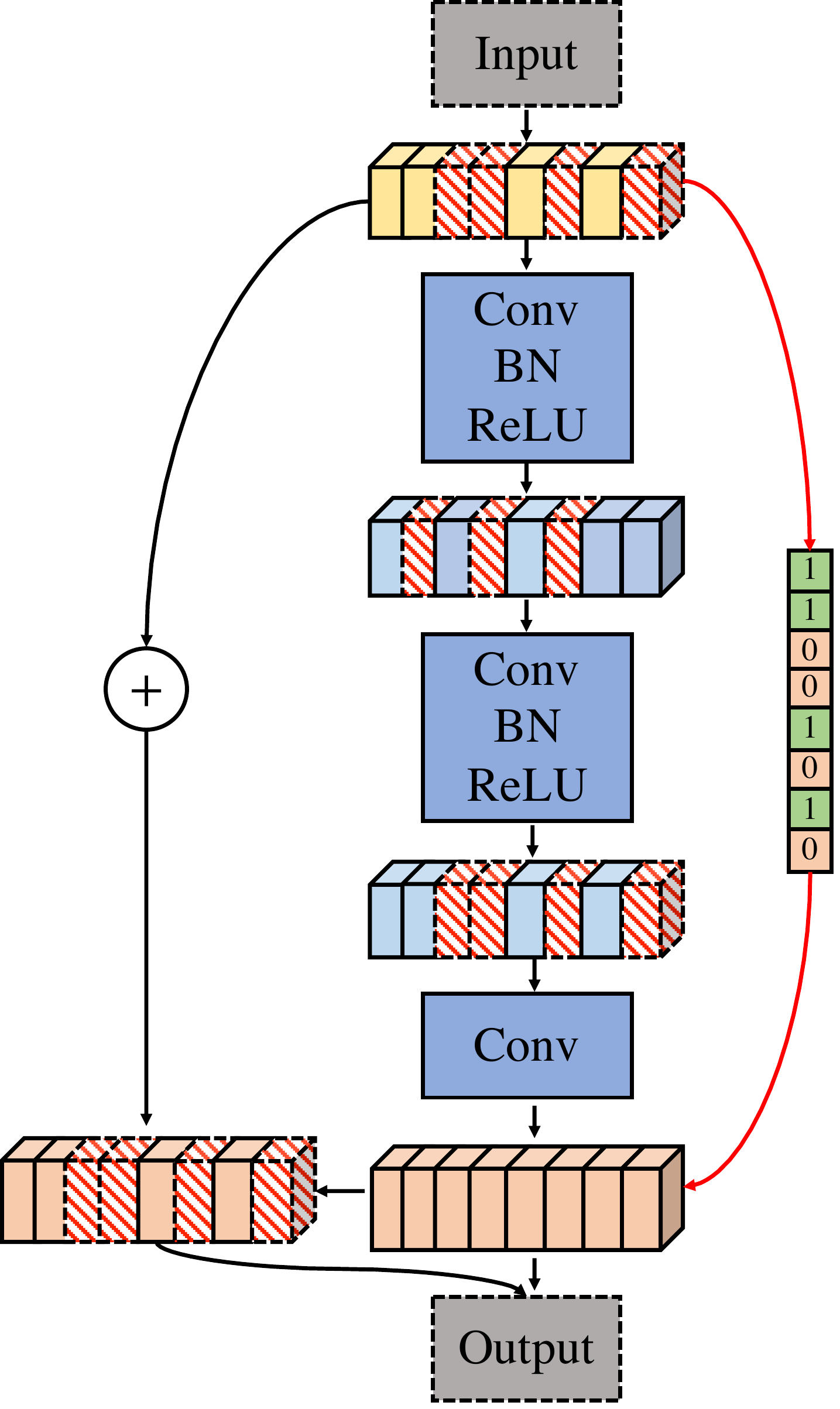}     
}   
\vspace{-3mm}
\caption{\small Zero-channel pruning for different modules.}     
\vspace{-6mm}
\label{fig:4}     
\end{figure}

\vspace{-4mm}
\section{Method}
\vspace{-4mm}
A universal style transfer algorithm usually consists of an auto-encoder and a feature transfer module. We propose a Zero-channel pruning algorithm and a sandwich swap feature transfer module to improve these two elements, respectively.
\vspace{-4mm}
\subsection{Zero-channel Pruning}
\vspace{-2mm}
To remove Zero Channels in features maps, we propose the \emph{Zero-channel Pruning} algorithm. Our algorithm prunes a network from the first layer to the last one.
Fig.~\ref{fig:4a} shows the way we prune a Conv-BN-ReLU module. We first store the indexes of every Zero Channel in a binary vector $m$. In $m$, every position is set to 1 or 0 to represent keeping/pruning the corresponding channel. For example, in Fig.~\ref{fig:4a}, the value of $m$ is set to ``11001010''. Then $m$ is passed to the upper convolutional and BN layers before the ReLU layer. Based on $m$, all Zero Channels in features and the corresponding weights in Conv and BN operators are removed.

To prune GoogLeNet, we specially adjust the proposed pruning method to deal with the Inception module. As Fig.~\ref{fig:4b} shows, we first prune every branch within the Inception module and then concatenate the compressed feature maps together. 
The proposed Zero-channel Pruning method also has the ability to prune other network architectures. To demonstrate its extensive applications, we also use it to prune Mobilenetv2.
Fig.~\ref{fig:4c} shows the way we prune Inverted Residual Blocks within Mobilenetv2. To enable the residual connections from the input to the output, we pass $m$ from the input feature to the output and prune it accordingly. By using the zero-channel pruning method, we reduce the parameter size of GoogLeNet from 6.63 MB to 3.28 MB and compress the Mobilenetv2 from 2.22 MB to 760.11 KB. Compared with other methods, our zero-channel pruning algorithm has two advantages. First, the pruned networks have comparable performance in making style transfer while achieving more than $2\times$ reduction of parameters compared with the original architectures. More importantly, our method achieves the above-mentioned results without any fine-tuning process.

 Fig.~\ref{fig:6} illustrates the network architecture we employ to conduct style transfer. The backbone network is based on the pruned GoogLeNet (ArtNet), we use the inverted version of the pruned GoogLeNet as the decoder to form an auto-encoder. 

\begin{figure}[t]
    \centering    
    \subfigure[\scriptsize{AdaIN}] {
 \label{fig:5a}     
\includegraphics[width=0.22\textwidth]{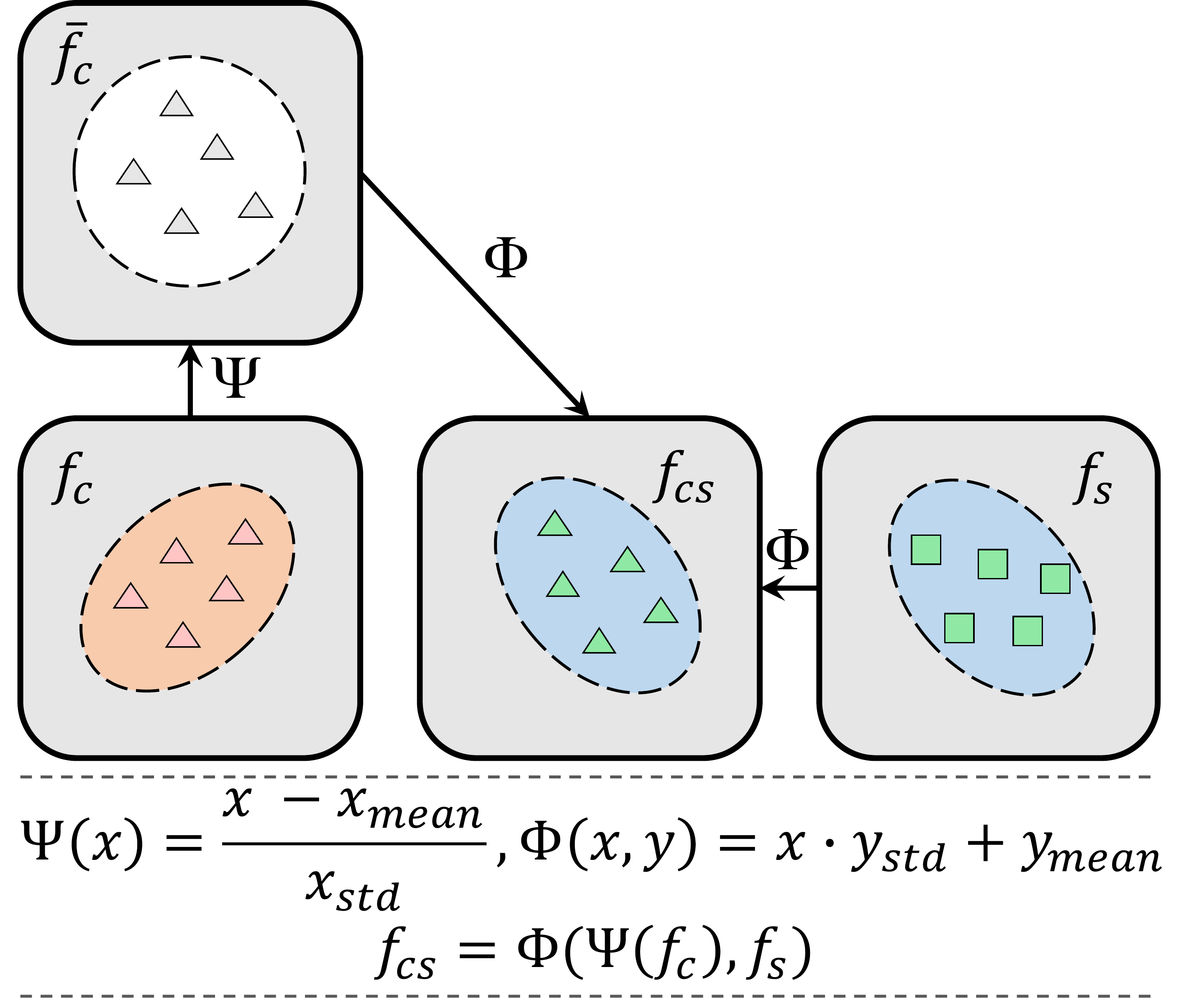}  
}      
\subfigure[\scriptsize{StyleSwap~\cite{chen2016fast}}] { 
\label{fig:5b}     
\includegraphics[width=0.22\textwidth]{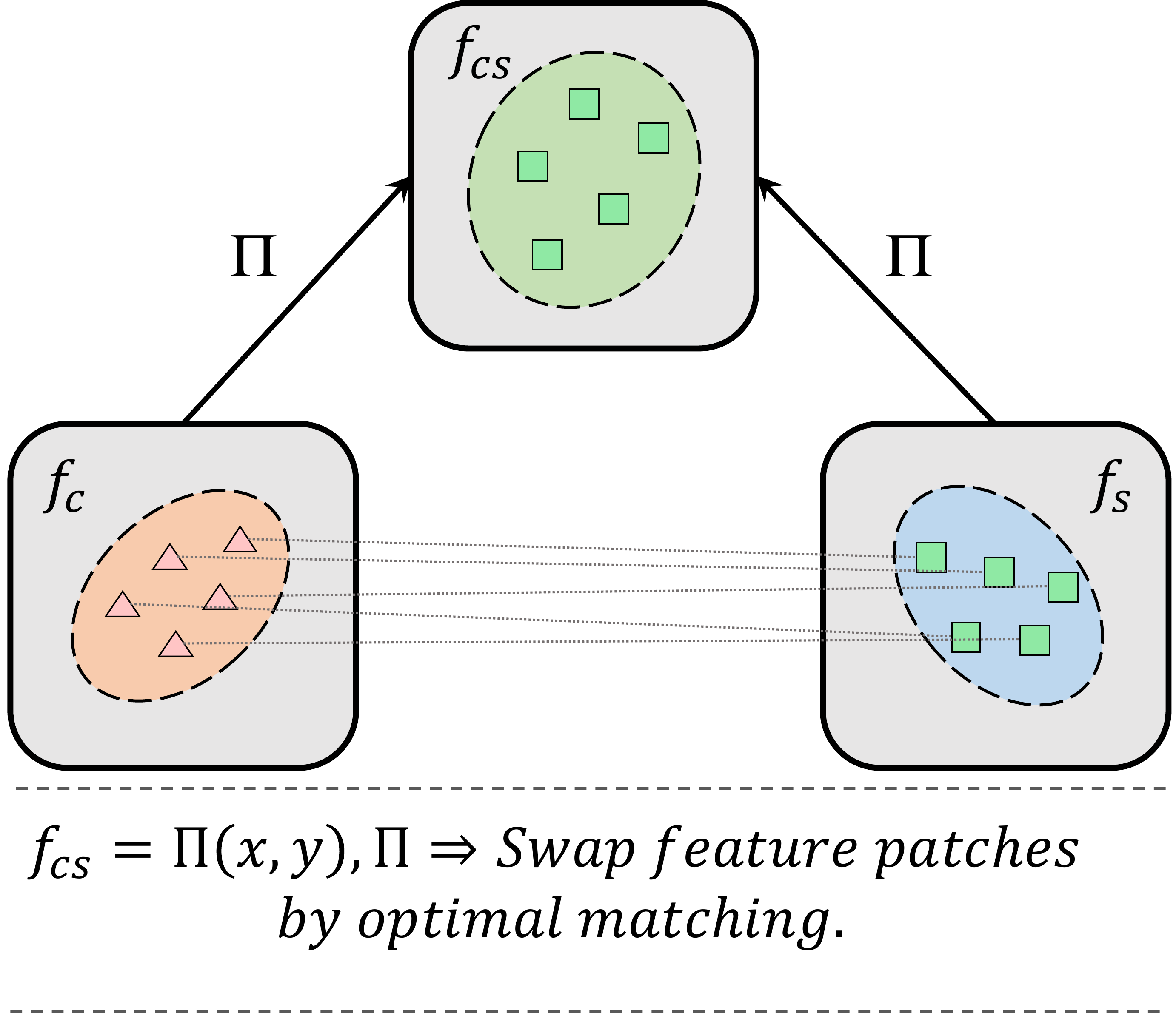}     
}    
\subfigure[\scriptsize{Style Decorator~\cite{sheng2018avatar}}] { 
\label{fig:5c}     
\includegraphics[width=0.22\textwidth]{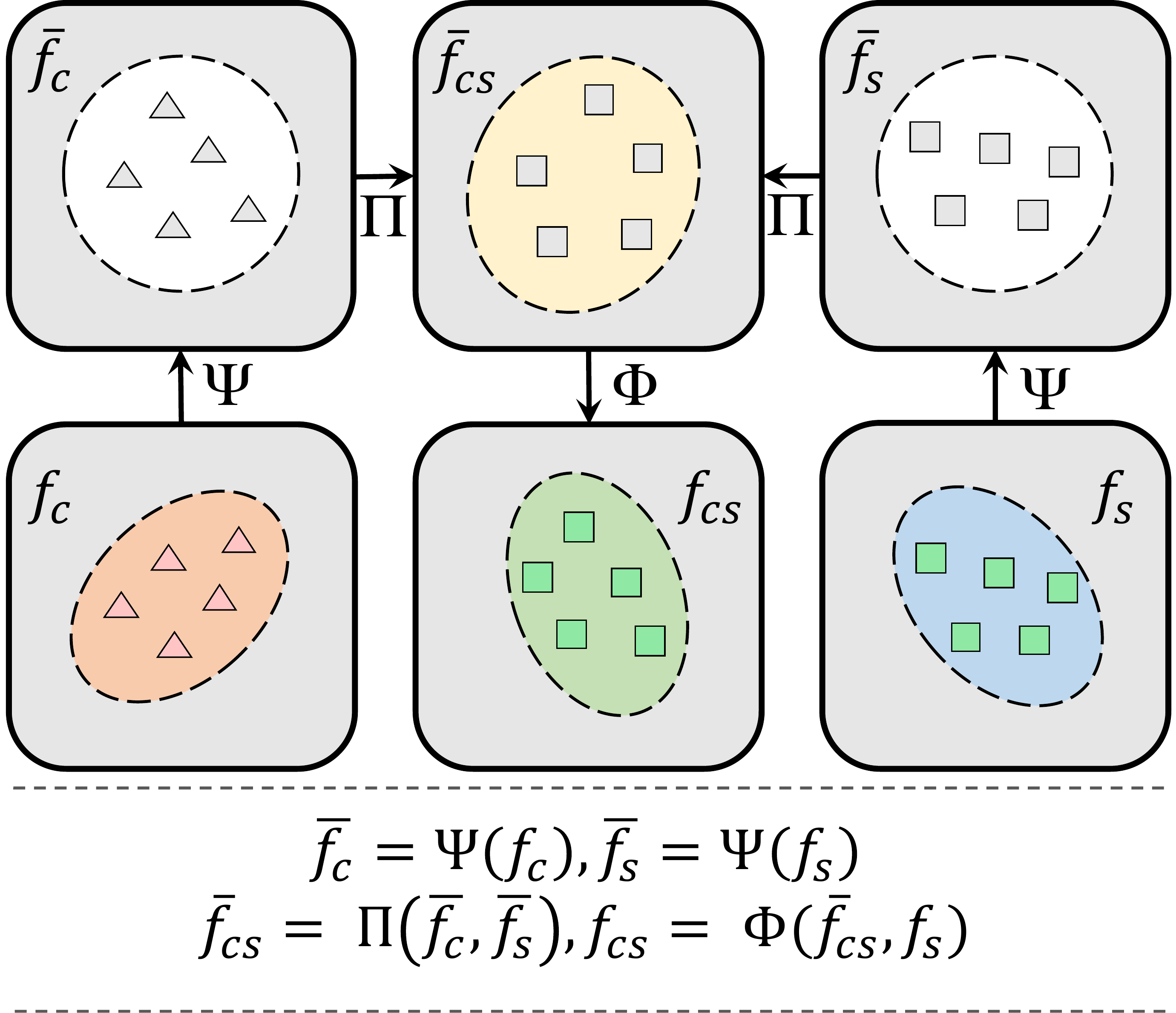}     
}   
\subfigure[\scriptsize{Ours $\mathbf{S}^2$}] { 
\label{fig:5d}     
\includegraphics[width=0.22\textwidth]{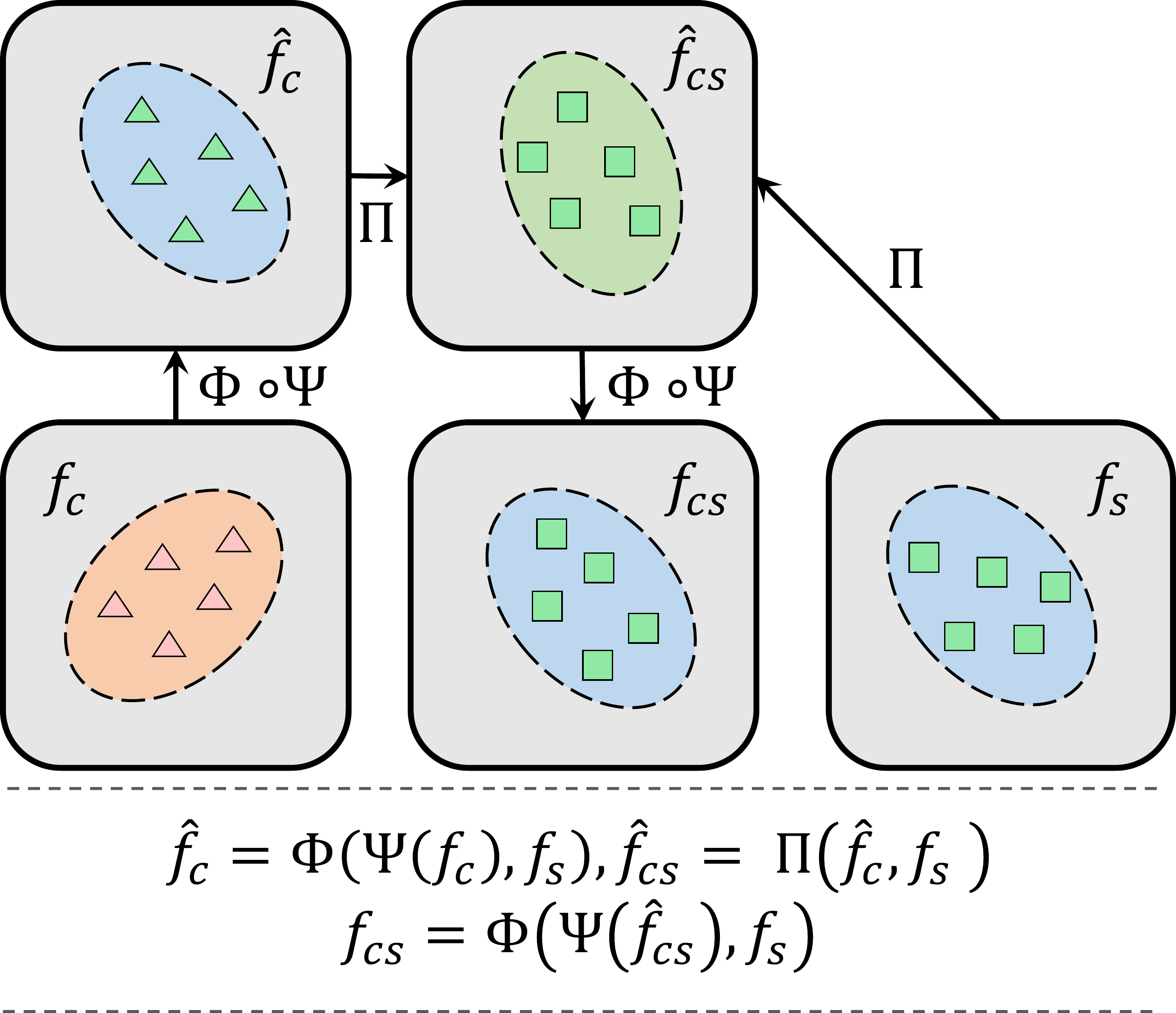}     
}   
\vspace{-2mm}
\caption{\small Comparison of different feature transfer modules. Shape of dotted ellipses means the variances of features, and the color denotes the means of features. The triangles and squares represent feature patches. (a) AdaIN transfers holistic global appearance by adopting $\Psi$ to normalize/whiten $f_c$ and then using $\Phi$ to match/coloring $\bar{f}_{cs}$ with respect to $f_s$. The produced $f_{cs}$ has the same mean $\mu$ and standard deviation $\sigma$ as $f_s$, but cannot transfer complex textures directly.  (b) StyleSwap creates $f_{cs}$ by using the $\Pi$ to get the optimal matching between $f_c$ and $f_s$. It has the ability to create fine textures but cannot transfer holistic global appearance since it cannot match $\mu$ and $\sigma$ between $f_{cs}$ and $f_s$. (c)  Avatar-Net also fails to match $\mu$ and $\sigma$, thus producing an impaired global appearance. Our $\mathbf{S}^2$ can benefit from $\Pi$ in rendering complicated textures and keep $\mu, \sigma$ of $f_{cs}$ equal to $f_s$.}     
\label{fig:5}     
\end{figure}
\begin{figure}[t]
    \centering
    \includegraphics[width=\textwidth]{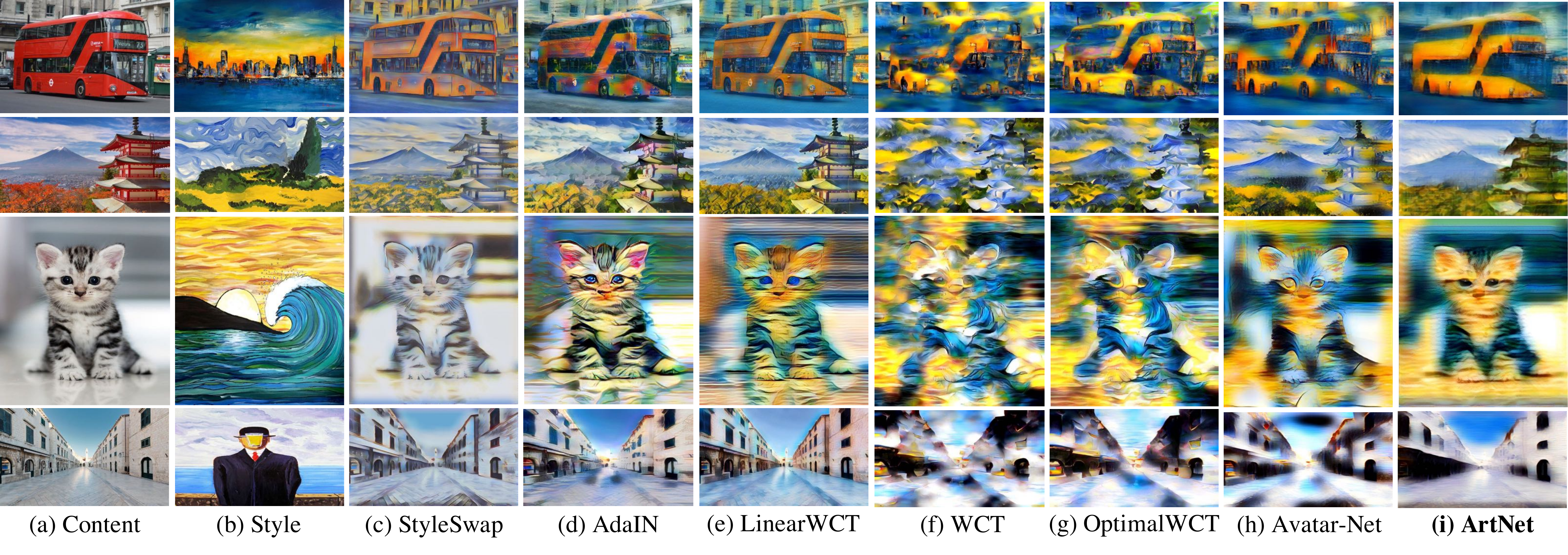}
    \vspace{-6mm}
    \caption{\small Style transfer result comparison against state-of-the-art universal style transfer algorithms. All compared images are generated by the officially released codes of the corresponding methods.}
    \vspace{-4mm}
    \label{fig:9}
\end{figure}
\vspace{-2mm}
\subsection{$\mathbf{S}^2$: Sandwich Swap Transform}
\vspace{-2mm}
To further improve style transfer effect, we propose a theoretically sound Sandwich Swap Transform ($\mathbf{S}^2$) module. $\mathbf{S}^2$ works at the bottleneck of ArtNet. As shown in Fig.~\ref{fig:6}, $\mathbf{S}^2$ adopts \emph{AdaIN-Swap-AdaIN} in a cascade to perform feature transfer. To design $\mathbf{S}^2$, we draw the inspiration from AdaIN~\cite{huang2017arbitrary} and StyleSwap~\cite{chen2016fast}. As Fig.~\ref{fig:5a} shows, AdaIN first normalizes the content feature $f_c$ with $\Psi$ and then re-colors it with respect to the style feature $f_s$ by $\Phi$. Specifically, AdaIN can maintain $\mu\left( f_{cs} \right) = \mu\left( f_s \right), \sigma\left( f_{cs} \right) = \sigma\left( f_s \right)$, where $\mu, \sigma$ represent the mean and standard deviation of the feature maps across $H, W$ axes and have a shape of $C\times1$. Note that the matching of $\mu$ and $\sigma$ can reduce the Gram loss between $f_{cs}$ and $f_s$, so AdaIN can transfer visual effects from the style to the content. However, AdaIN can only transfer global appearance and usually fails in rendering complicated textures. StyleSwap (Fig.~\ref{fig:5b}) directly borrows fine textures from $f_s$ to $f_{cs}$ according to the optimal matching between $f_c$ and $f_s$. Since $f_{cs}$ consists of a few selected patches of $f_s$, $\mu$ and $\sigma$ of $f_{cs}$ and $f_s$ are different. Moreover, directly making swap on unnormalized $f_c$ and $f_s$ may bias the optimal matching~\cite{sheng2018avatar}. Therefore, the result of StyleSwap usually contains significant color aberrations and artifacts. Our $\mathbf{S}^2$ combines the advantages of both AdaIN and StyleSwap. More specifically, the $\mathbf{S}^2$ module (Fig.~\ref{fig:5d}) first adopts an AdaIN module to project $f_c$ to the space of $f_s$, and then a swap module is used to directly copy fine textures/painting strokes from $f_s$ to $\hat{f}_{cs}$. After that, another AdaIN module is introduced to correct color aberrations by aligning $\mu$ and $\sigma$ between $f_{cs}$ and $f_s$. 

Compared with Avatar-Net, 
our method makes style swap in the space of $f_s$ instead of the normalized space. In this way, the style swap procedure can achieve a more accurate patch matching by taking both textures and colors to perform optimal matching while colors are not considered in Avatar-Net. Based on it, our projection can achieve inter-object distinction and inner-object color consistency in style transfer, thus facilitating content preservation.
On the other hand, in Avatar-Net, since $\mu\left( f_{cs} \right) \ne \mu\left( f_{s} \right), \sigma\left( f_{cs} \right) \ne \sigma\left( f_{s} \right)$, the reconstructed image directly from $f_{cs}$ has compromised stylization effect. To overcome this problem, Avatar-Net uses a ``Hourglass'' strategy as a remedy. In contrast, we introduce an additional normalization step $\Psi$ before the recoloring procedure $\Phi$ to force $\mu$ and $\sigma$ of $f_{cs}$ to be the same as those in $f_s$ (Fig.~\ref{fig:5d}), therefore leading to a reduced Gram loss and an improved style transfer effect. Interestingly,  $\Psi \circ \Phi$ forms a new AdaIN module ($i.e.$, the last AdaIN in $\mathbf{S}^2$). Inspired by An~\etal\cite{an2019ultrafast}, we concatenate deep features from all encoder blocks together and feed the concatenated $f_c$ and $f_s$ into $\mathbf{S}^2$ (Fig.~\ref{fig:6}). This strategy enables $\mathbf{S}^2$ to make use of the features from high level to low level without introducing an additional computational burden. (Fig.~\ref{fig:8}(h) shows the ablation result without this strategy.) Thanks to the flexibility of $\mathbf{S}^2$, our method allows transfer textures from a reference image and colors from another one separately. Please refer to the supplementary material for details.

\vspace{-3mm}
\section{Experiments}
\vspace{-3mm}
In this section, we present evaluation results against the state-of-the-art style transfer methods. 
Detailed experimental settings and more results can be found in the supplementary material.

\begin{table}[t]
    \caption{\small Quantitative evaluation results for universal stylization methods. Higher is better.}
\centering
    \scriptsize 
     \tabcolsep=0.295cm
    \begin{tabular}{lcccccccc}
        \toprule
        Method & StyleSwap & AdaIN & WCT & LinearWCT & OptimalWCT & Avatar-Net & \textbf{ArtNet} \\
        \midrule
        SSIM~$\uparrow$ & 0.4851 & 0.3525 & 0.2032 & 0.4363 & 0.2511 & 0.3829 & \textbf{0.4452}\\
        User Preference (\%)~$\uparrow$ & 7.38 & 8.20 & 3.28 & 26.50 & 4.37 & 14.48 & \textbf{35.79}\\
        \bottomrule
    \end{tabular}
    \vspace{-5mm}
    \label{tab:3}
\end{table}
\begin{table}[t]
    \caption{\small Computing-time comparison. OOM means out of the memory and NA means that the code cannot run at this resolution.}
    \centering
    \scriptsize \setlength{\tabcolsep}{9.7pt}
    \begin{tabular}{lcccccccc}
        \toprule
        Method & StyleSwap & AdaIN & WCT & LinearWCT & OptimalWCT & Avatar-Net & \textbf{ArtNet}\\
        \midrule
        $128\times128$ & 0.0478 & 0.0037 & 2.6873 & 0.0051 & 0.5003 & NA & \textbf{0.0142}\\
        $256\times256$ & 0.3068 & 0.0093 & 3.0805 & 0.0167 & 0.8793 & 0.1732 & \textbf{0.0145}\\
        $512\times512$ & 1.5782 & 0.0344 & 4.1922 & 0.0603 & 1.8077 & 0.3718 & \textbf{0.0147}\\
        $1024\times1024$ & OOM & 0.1363 & OOM & 0.2278 & 4.1589 & OOM & \textbf{0.0775}\\
\bottomrule
    \end{tabular}
    \vspace{-2mm}
\label{tab:4}
\end{table}

\vspace{-3mm}
\subsection{Visual Comparison}
\vspace{-2mm}
We evaluate the effectiveness of our method in comparison to state-of-the-art universal methods: StyleSwap~\cite{chen2016fast}, AdaIN~\cite{huang2017arbitrary}, WCT~\cite{li2017universal}, Avatar-Net~\cite{sheng2018avatar}, LinearWCT~\cite{li2018learning} and OptimalWCT~\cite{lu2019optimal}.

StyleSwap 
fails to transfer complicated textures and colors as shown in Fig.~\ref{fig:9}(c). AdaIN (d) and LinearWCT (e) can generate complex details. However, the generated images are visually dissimilar to the corresponding style images in terms of colors (row 1/3) and local textures (row 2/4). WCT (f) and OptimalWCT (g) are good at creating visually pleasing local textures and bright colors. However, this approach puts too much emphasis on local details and makes the generated images fragmented such that the content is unrecognizable. Avatar-Net (h) improves the results of WCT in terms of preserving characters and objects in the content images. However, the transferred images are distorted. For example, in the first row, Avatar-Net paints the bus body with inconsistent blue and orange colors, but the color of the bus in the content image is uniform. Furthermore, in the second and fourth rows, Avatar-Net renders the sky in images with the colors of the land (row 2) and windows (row 4), causing severe artifacts. With faithful color matching and high-quality texture generation, our method (i) outperforms the compared algorithms in terms of ensuring uniform stylization effect within every object of an image, and therefore has far fewer artifacts and distortions. For example, the bus and the street (row 1), the sky and the land (row 2) have distinct styles, but the visual effects within each object are consistent. This advantage enables a much stronger ability of our method to faithfully preserve rich information from content pictures while maintaining a high-quality style transfer.

\vspace{-3mm}
\subsection{Quantitative and Qualitative Comparisons}
\vspace{-2mm}
To demonstrate the effectiveness of the proposed method, we conduct a quantitative comparison against state-of-the-art methods upon a dataset that consists of 1092 content-style pairs (42 content; 26 style). All images are acquired from the Internet. Inspired by~\cite{yoo2019photorealistic}, we adopt the Structural Similarity Index (SSIM) between edge responses~\cite{xie2015holistically} of original contents and stylized images as the metric to measure the performance of the content information preservation in style transfer. Since the objective evaluation of style transfer effect remains an open problem, we conduct a user study to subjectively assess the stylization effect by all the compared methods. We show the quantitative evaluation results for universal stylization methods in Table~\ref{tab:3}. StyleSwap has the highest SSIM score but the lowest user preference. This is because the results of StyleSwap are extremely biased toward content preservation. The proposed ArtNet has the highest user preference while its SSIM score ranks only second to StyleSwap, suggesting that our algorithm can achieve high-quality style transfer results and faithful content preservation simultaneously. 

\vspace{-3mm}
\subsection{Computational Time Comparison}
\vspace{-2mm}
We conduct a computing time comparison against the state-of-the-art universal methods to demonstrate the real-time efficiency of the proposed network architectures.
All approaches are tested on the same computing platform which includes an NVIDIA TitanXp GPU card with 16GB RAM.
We compare the computing time on content and style images of different resolutions. As Table~\ref{tab:4} shows, the proposed ArtNet outperforms the compared methods in terms of efficiency. ArtNet can achieve real-time (around 68 FPS) style transfer at the resolution of $512\times512$.

\vspace{-3mm}
\subsection{Ablation Study}
\begin{figure}[t]
    \centering
    \includegraphics[width=\textwidth]{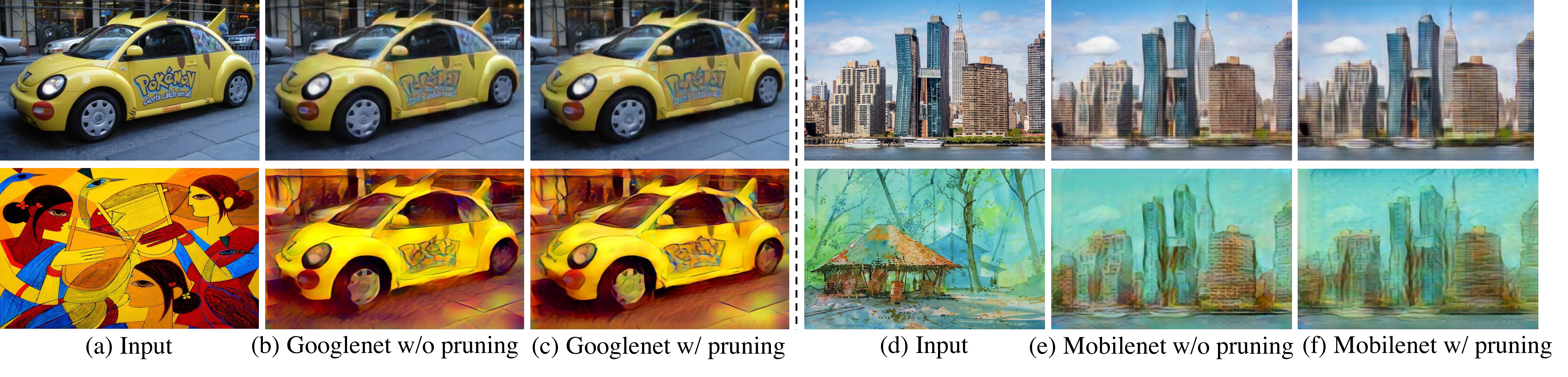}
\vspace{-6mm}
    \caption{\small Comparison between the style transfer results with/without pruning. Top row: Image reconstruction results. Bottom row: Style transfer results.}
\vspace{-4mm}
    \label{fig:7}
\end{figure}

\vspace{-1mm}
\paragraph{Zero-channel Pruning.}
\vspace{-2mm}
In Fig.~\ref{fig:7}, we show the image reconstruction (top row) and style transfer (bottom row) results by using GoogLeNet \emph{v.s.} pruned GoogLeNet (c) and Mobilenetv2 (e) \emph{v.s.} pruned Mobilenetv2 (f). As shown in the top row of Fig.~\ref{fig:7}, the zero-channel pruning algorithm does not harm the image reconstruction ability of the auto-encoder. Please refer to the supplementary material for the quantitative comparison between the pruned and unpruned networks. More importantly, the style transfer results with/without the proposed pruning algorithm are almost the same, which shows that the zero-channel pruning method can reduce the parameter size of the network without hurting the quality of style transfer results.

\vspace{-1mm}
\paragraph{Sandwich Swap Module.}
\vspace{-2mm}
\begin{figure}[t]
    \centering
    \includegraphics[width=\textwidth]{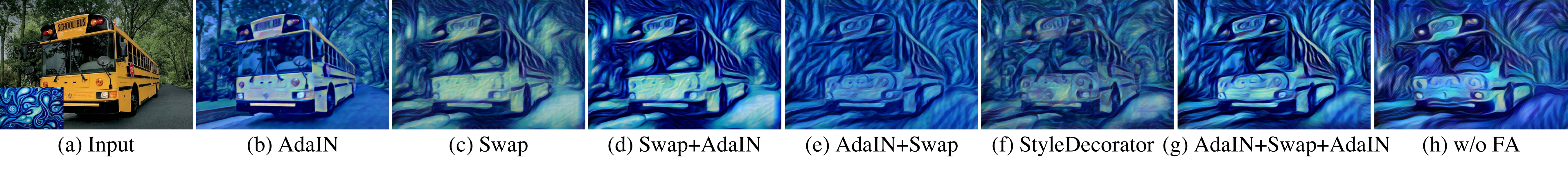}
\vspace{-6mm}
    \caption{\small Ablation study for the proposed $\mathbf{S}^2$ module and feature aggregation. Please compare the transfer module by Avatar-Net (f) and ours (g).}
\vspace{-6mm}
    \label{fig:8}
\end{figure}
To demonstrate the effectiveness of the proposed $\mathbf{S}^2$ module, we conduct an ablation study to each element of it. As shown in Fig.~\ref{fig:8}(b), the AdaIN~\cite{huang2017arbitrary} module can transfer holistic global appearance. However, it cannot preserve fine textures. Fig.~\ref{fig:8}(c) shows the style transfer results only with the style swap module~\cite{chen2016fast}. Although the results contain rich details, the generated image contains a nasty aberration in terms of color matching. 
The prior AdaIN in $\mathbf{S}^2$ is to project $f_c$ into the domain of $f_s$, and thus correct the bias of the optimal matching (Fig.~\ref{fig:8}(e)). In contrast, the posterior AdaIN can rematch the mean and variance of the output feature after the style swapping to $f_s$, which can correct the color aberration introduced by the style swap module (Fig.~\ref{fig:8}(d)). Fig.~\ref{fig:8}(g) shows the style transfer results by $\mathbf{S}^2$. Our algorithm, with improved content preservation, can achieve exact color matching and rich-detail rendering at the same time.

\vspace{-3mm}
\section{Conclusion}
\vspace{-3mm}
In this paper, we proposed the first style transfer algorithm that can achieve universal, real-time, high-quality style transfer simultaneously. Our method adopts a light-weight auto-encoder based on GoogLeNet. 
Furthermore, we presented a Zero-channel Pruning method to remove redundant Zero Channels in feature maps produced by extensive networks. Via zero-channel pruning, GoogLeNet can achieve real-time efficiency in style transfer on high-resolution images. Moreover, we introduced a sandwich swap transform module to transfer artistic styles at the bottleneck of the auto-encoder. With the proposed modules, the proposed transfer method achieves real-time, high-quality stylization, and faithful content preservation at the same time. The extensive experiments
 show that the proposed algorithm can remarkably improve the stylization effect and the content preservation ability, while reducing the computational overhead dramatically. 

\clearpage

\small
\bibliographystyle{plain}
\bibliography{ms}

\end{document}